\documentclass{article}

\usepackage{microtype}
\usepackage{graphicx}
\usepackage{subfigure}
\usepackage{tabularx}
\usepackage{color, colortbl}
\usepackage{xcolor}
\usepackage{booktabs} 
\usepackage{enumitem}
\usepackage{hyperref}

\usepackage[accepted]{icml2024}

\usepackage{amsmath}
\usepackage{amssymb}
\usepackage{mathtools}
\usepackage{amsthm}

\usepackage[capitalize,noabbrev]{cleveref}

\theoremstyle{plain}

\theoremstyle{definition}

\theoremstyle{remark}

\usepackage[textsize=tiny]{todonotes}
\usepackage{multirow}

\icmltitlerunning{Characterizing Truthfulness in Large Language Model Generations with Local Intrinsic Dimension}

\begin{document}

\twocolumn[
\icmltitle{Characterizing Truthfulness in Large Language Model Generations \\ with Local Intrinsic Dimension}




\begin{icmlauthorlist}
\icmlauthor{Fan Yin}{yyy}
\icmlauthor{Jayanth Srinivasa}{comp}
\icmlauthor{Kai-Wei Chang}{yyy}
\end{icmlauthorlist}

\icmlaffiliation{yyy}{Department of Computer Science, University of California, Los Angeles, LA, U.S.A.}
\icmlaffiliation{comp}{Cisco Research, U.S.A}

\icmlcorrespondingauthor{Fan Yin}{fanyin20@cs.ucla.edu}


\vskip 0.3in
]



\printAffiliationsAndNotice{}  

\begin{abstract}
We study how to characterize and predict the truthfulness of texts generated from large language models (LLMs), which serves as a crucial step in building trust between humans and LLMs. Although several approaches based on entropy or verbalized uncertainty have been proposed to calibrate model predictions, these methods are often intractable, sensitive to hyperparameters, and less reliable when applied in generative tasks with LLMs. In this paper, we suggest investigating internal activations and quantifying LLM's truthfulness using the local intrinsic dimension (LID) of model activations. Through experiments on four question answering (QA) datasets, we demonstrate the effectiveness of our proposed method. Additionally, we study intrinsic dimensions in LLMs and their relations with model layers, autoregressive language modeling, and the training of LLMs, revealing that intrinsic dimensions can be a powerful approach to understanding LLMs. Code is available at: \url{https://github.com/fanyin3639/LID-HallucinationDetection}.

\end{abstract}

\section{Introduction}
\label{intro}
nLarge language models (LLMs) have demonstrated remarkable effectiveness in various generative natural language processing (NLP) tasks, including QA, summarization, and dialogue~\citep{touvron2023llama, chowdhery2023palm, OpenAI_GPT4_2023}. However, deploying LLMs to more high-stakes scenarios remains limited due to their tendency to provide plausible but untruthful answers, even when they are uncertain, also known as hallucinations~\citep{ji2023survey}. Hence, characterizing and eliciting the truthfulness of model outputs is a crucial step towards constructing more reliable LLMs and building user trust in models~\citep{bommasani2021opportunities, kadavath2022language, zou2023representation}.

\begin{figure}
    \centering
    \includegraphics[scale=0.4]{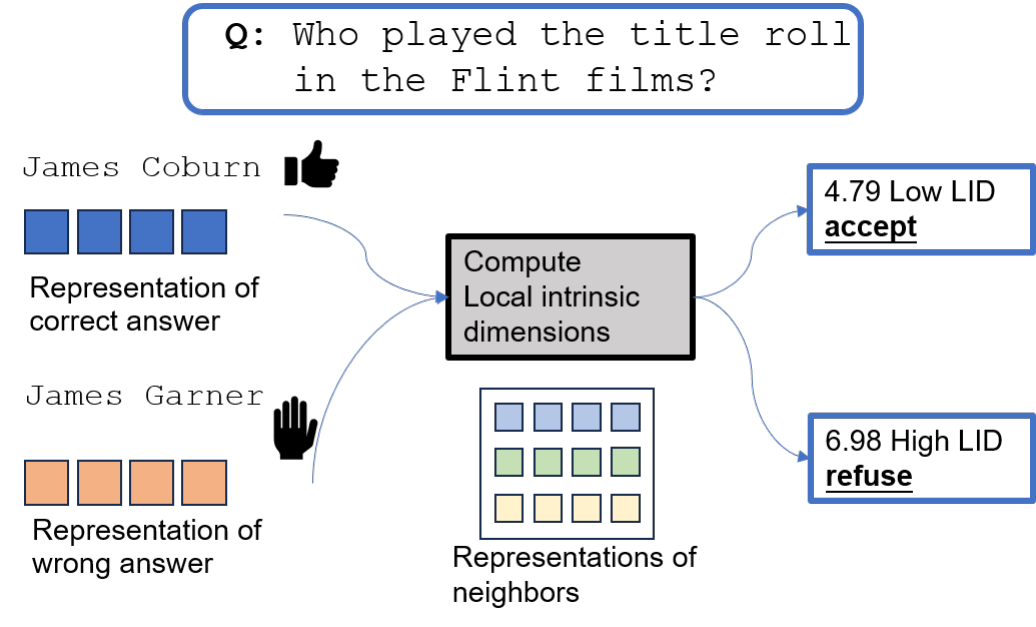}
    \caption{Detecting hallucinations with LIDs. LLM representations of correct answers have smaller intrinsic dimensions.}
    \label{fig:pipline}
\end{figure}
Despite its importance, little is known about which information within models most accurately characterizes their truthfulness. A mainstream of work approaches this through logit-level entropy-based uncertainty~\citep{gal2016dropout, malinin2020uncertainty, kuhn2022semantic, duan2023shifting} or verbalized uncertainty~\citep{kadavath2022language, zhou2023navigating, tian2023just}. However, computing uncertainty is limited to classification tasks and becomes intractable for generative tasks due to infinite output space. Moreover, extracting truthfulness only at the output layer inevitably loses substantial information, leading to sub-optimal performance.


Other approaches train linear probes to discover truthfulness directions in model internal representations~\citep{azaria2023internal, li2023inference, burns2022discovering, zou2023representation, marks2023geometry}. However, these truthful directions do not always exist and can vary significantly due to tasks, the layers being used, and the styles of prompts. Therefore, determining whether a direction would be beneficial for the task at hand can be cumbersome.
 
In this paper, we delve into the internal representations, which have been shown to preserve more information and geometric characteristics. Instead of seeking truthful directions for each task, we leverage a more principled and generalizable feature to detect hallucinations: the discrepancy in local intrinsic dimension (LID)~\citep{levina2004maximum, pmlr-v101-gomtsyan19a} of model activations. LID reflects the minimal number of activations required to characterize the current point without significant information loss. A higher LID means that the current point lies in a more complicated manifold and vice verse. LLM representations, which are often high-dimensional vectors (e.g., 4,096 for Llama-2-7B,~\citealt{touvron2023llama2}) are commonly believed to lie in lower-dimensional manifolds because of the inductive bias of the model and the natural structure in human language~\citep{marks2023geometry}. We hypothesize that truthful outputs, being closer to natural language, are more structured and have smaller LIDs. On the other hand, an untruthful continuation of a prompt hallucinated by the model itself would mix human (prompt) and complex model (continuation) distribution, leading to larger LIDs. The discrepancy in LID would thus serve as a strong signal to assess whether an output is truthful or not.


More specifically, our method is based on the well-established maximum likelihood estimation (MLE) method ~\citep{levina2004maximum} but proposes a simple yet effective correction to 1) accommodate the non-linearity in language representations~\citep{pmlr-v101-gomtsyan19a}; 2) select the optimal set of representations. MLE approximates the count of neighbors surrounding the current sample with a Poisson process parameterized by the LID. It inherently supports estimation for an individual sample while other estimators mostly consider intrinsic dimension as a property of the whole dataset. Our improvements enable more accurate estimations of LIDs in representations. 

Experiments with the Llama-2~\citep{touvron2023llama2} family on four QA tasks prove the advantage of using LID methods over uncertainty methods, achieving an improvement of 8\% under AUROC. Compared with representation-level methods like linear probes and t-SNE~\citep{van2008visualizing}, we show that our method is more powerful while other methods fail to discover truthful directions.  Further ablation study shows that we could even leverage out-of-distribution samples as neighbors to estimate the LIDs, demonstrating the generalizability of our methods.

We further conduct a series of analyses on the intrinsic dimensions in LLMs, revealing several intriguing properties beyond its relations to hallucinations. Overall, we believe intrinsic dimension is an insightful and powerful feature for understanding LLMs. Below are our findings:
\begin{itemize}[itemsep=0.5mm, parsep=0pt, leftmargin=*]
    \item  Similar to the findings by~\citet{ansuini2019intrinsic} on image data, we observe a `hunchback' shape in the intrinsic dimension of language generations: the intrinsic dimension values increase in the first few layers and then gradually decrease. The hallucination detection performance curve follows a similar shape to the intrinsic dimension value curve but is `shifted behind' by one or two layers;
    \item  We verify our hypothesis that mixing human and model distributions increases intrinsic dimension by controlling where the `mixing' happens. We show that the intrinsic dimensions for human answers are consistently lower than untruthful model outputs at every position, exhibiting a sharp decrease when the answer approaches the end;
    \item  In addition to frozen foundation model, we are curious about how instruction tuning, a technique widely adopted to align LLMs, impacts intrinsic dimensions. We find that as the instruction tuning progresses, the intrinsic dimension of LLMs' representations tends to increase. Furthermore, intrinsic dimensions correlate with the generalization performance of the model.
\end{itemize}
\section{Related Work}
\noindent{\bf Characterizing Model Truthfulness} As LLMs improve, it is increasingly crucial to ensure their safety and truthfulness~\citep{hendrycks2021unsolved, bommasani2021opportunities}. An important technique is to detect incorrect model outputs so that users can decide when not to 
 trust those outputs~\citep{kamath2020selective, ren2022out}. Existing techniques towards this goal mainly fall into three categories: 1) entropy-based uncertainty estimation~\citep{malinin2020uncertainty, kuhn2022semantic, duan2023shifting, lin2023generating}. However, those approximations are not accurate for LLMs since the output space of LLMs is too large; 2) Verbalized uncertainty~\citep{kadavath2022language, tian2023just, zhou2023navigating, xiong2023can}, i.e., directly asking LLMs to judge their answers. 
It typically involves extra training as models are not pre-trained with this objective; 3) Probing truthfulness direction~\citep{zou2023representation, azaria2023internal, li2023inference, burns2022discovering} that tries to find a truthfulness direction in model representations for a specific dataset. The obtained direction usually does not generalize well. We propose to use LIDs to detect hallucinations, which leverage geometric information that uncertainty-based methods ignore and are more generalizable than truthful directions.

\noindent{\bf (Local) Intrinsic Dimension in Neural Models} Neural models are commonly believed to be redundant in terms of their parameters and representations~\citep{birdal2021intrinsic}. Approaches to estimate intrinsic dimension of data manifolds~\citep{levina2004maximum, amsaleg2015estimating, facco2017estimating} have been applied to understand the structure in models. For example,~\citet{ma2018characterizing} use differences in LID values to characterize adversarial image data;~\citet{ansuini2019intrinsic, pope2020intrinsic, birdal2021intrinsic} study the relation between intrinsic dimensions and generalization ability of models. Most related to ours,~\citet{tulchinskii2023intrinsic} train classifiers using ID values to identify AI-generated texts in multiple languages. Different from previous work, we focus more on LID for each individual sample, and use LID to identify incorrect model outputs.

\section{LID for Characterizing Truthfulness}
In this section, we formulate the problem of characterizing the truthfulness of model outputs. We review the MLE framework for LID and introduce the modifications to account for the LLMs' representations.
\subsection{Problem Setup}
Consider an $L$-layer causal LLM $M$ that takes a sequence of $N$ tokens $X=\left[x_1, x_2, \dots, x_N\right]$ as input, and generates a sequence of $O$-token continuation as output $M\left(X\right) = \left[x_{N + 1}, x_{N + 2}, \dots, x_{N + O}\right]$. $M\left(X\right)$ is generated in an autoregressive manner, where each $x_{N + i}, i \in \left[1, \dots, O\right]$ is sampled from a distribution over the model vocabulary $\mathcal{V}$, conditioned on the prefix $\left[x_1, x_2, \dots, x_N, x_{N + 1}, \dots, x_{N + i-1}\right]$:
$$
     \mathbf{X}_{Li} = \left( M_{L} \circ M_{L-1} \circ \dots \circ M_0 \right) \left( \left[ x_1 , \dots, x_{N + i-1} \right] \right),
$$
$$
    p\left(x_{N + i} | \left[ x_1 \dots, x_{N + i-1} \right]\right) = \text{softmax} \left( W \mathbf{X}_{Li} + b \right),
$$
where $M_j,\,j \in \left[1 \dots L\right]$ is the j-th layer of the LLM $M$. $M_0$ is the embedding layer. $W, b$ are the output projection weights and bias. For later sections, we use $\mathbf{X}_{ji} \in \mathcal{R}^D$ to denote the j-th layer representation for the i-th continuation token $x_{N + i}$, which is a $D$-dimensional vector. We denote the probability distribution over a single token $x_{N+i}$ as $p(x_{N + i} | \cdot)$, and the whole sequential output $M(X)$  $p(M(X) | \cdot)$, respectively.

For a specific task with $n$ points $\mathcal{D} = \left\{X^1, \dots, X^n\right\}$, we aim to predict the truthfulness of each corresponding generation $\left\{M(X^1) \dots M(X^n)\right\}$ before knowing the ground truth. Note that the truthfulness criteria might vary based on the task being considered, which might be string matching, semantic similarity, or any other human-based metrics. We use $\left\{\hat{Y}^1 \dots \hat{Y}^n\right\}$ to denote the ground truth, and $s\left(M(X^i),\,\hat{Y}^i\right) \in \{0, 1\}$ to denote the indicator function for whether an input-output pair ($X^i$, $M(X^i)$) can be considered truthful. The goal of this paper is to propose a characterizing feature that accurately reflects $s\left(M(X^i),\,\hat{Y}^i\right)$. While previous works on uncertainty estimation mostly obtain the feature from the final predictive distribution $p(M(X) | \cdot)$, we instead propose to explore the LID of intermediate representations $\mathbf{X}_{ji}$.

\subsection{MLE Estimator for LID}
Here, we review the core idea of the MLE estimator for LIDs~\citep{levina2004maximum}. Notice that while there are other estimators available for intrinsic dimension~\citep{Costa2005EstimatingLI, Facco2017EstimatingTI, campadelli2015intrinsic}, MLE is specifically tailored for estimating `local' intrinsic dimension of an individual point, making it well-suited for our application, unlike many others that estimate `global' intrinsic dimension.

For the representation\footnote{We use the representation from a middle layer and the last token, elaborated in Section\ref{section:methodls}. For simplicity, we omit the subscripts for its position and layer.} of a data point in $\mathcal{D}$, $\mathbf{X}^i$, the MLE estimator fits a Poisson process to the count of the neighbors around where the rate of the Poisson process is parameterized by the intrinsic dimension $m$. Formally, it considers the $T$ nearest neighbors of $\mathbf{X}^i$ in $\mathcal{D}$, $\left\{\mathbf{X}^{i1}, \dots, \mathbf{X}^{iT}\right\}$, and a ball of radius $R$ centered at $\mathbf{X}^i$, $S_{\mathbf{X}^i}(R)$. The count of neighbors inside balls of varying radius $0 < t < R$ can be expressed by a binomial process as follows:
$$
N\left(t,\,\mathbf{X}^{i}\right) = \sum_{k=1}^{T} \mathbb{I}\left\{\mathbf{X}^{ik} \in S_{\mathbf{X}^i}(t)\right\}.
$$
\citet{levina2004maximum} propose to approximate the above with a Poisson process of a certain rate $\lambda\left(t\right)$. According to the definition of $\lambda$, if we assume the density $f$ is approximately constant around $\mathbf{X}^i$, and the volume $V$ expands proportionally to $t^m$, i.e., $V = V_m t^m$ where $m$ is the intrinsic dimension, we will have 

$$\lambda\left( t \right) = f \frac{dV}{dt} = f V_m\,m\,t^{m-1}.$$ 

Then, the log-likelihood of the Poisson process can be written as a function of the intrinsic dimension $m$ and $\theta = log\,f$.
\begin{equation}
   L\left(m, \theta \right) = \int_{0}^R log \lambda\left(t\right) d N_{\lambda} \left(t\right) - \int_0^{R} \lambda\left(t\right) dt.
\label{eq:likelihood}
\end{equation}
Maximizing the log-likelihood in Eq. \ref{eq:likelihood}, we have the following formula for calculating the intrinsic dimension $m$:
\begin{equation}
   m\left(R, \mathbf{X}^i\right) = \left( \frac{1}{N(R, \mathbf{X}^i)} \sum_{j=1}^{N(R, \mathbf{X}^i)}log\frac{R}{Q_j}\right) ^{-1},
\label{eq:IC}
\end{equation}
where $Q_j,\,j=1, \dots T$ is the Euclidean distance to the j-th nearest neighbor. The numerical calculation of Eq. \ref{eq:IC} can be further simplified as:
\begin{equation}
   m\left(\textbf{X}^i\right) = \left( \frac{1}{T - 1} \sum_{j=1}^{T-1}log\frac{Q_{T}}{Q_j}\right) ^{-1}.
\label{eq:IC_num}
\end{equation}

\subsection{Layer Selection and Distance-aware MLE}
\label{section:methodls}
In the previous section, we discuss how to calculate LIDs for a representation $\textbf{X}^i$. When it comes to LLMs, two challenges arise: 1) there will be a $D$-dimensional representation for each position at each layer, making it hard to select the optimal representation to use; 2) MLE assumes constant density function $f$, which is unlikely to hold for causal LLMs on complicated real data. Next, we discuss our solutions for addressing those issues.

\noindent{\bf Layer Selection} As mentioned earlier, LLMs generate a $D$-dimensional representation for each token at each layer. We select the token at the last position of $\mathbf{X}^i$, i.e., $\mathbf{X}_{-1}^i$ as that representation contains all pertinent information from preceding positions. This strategy aligns with other works on probing truthful directions like ~\citet{zou2023representation}. 

However, when it comes to layer selection, our empirical evidence indicates that the representations from the last layer might not yield the most informative feature. As in Figure \ref{fig:delaycurve}, the performance of predicting truthfulness with LIDs correlates well with the absolute value of summed LIDs over the test set, but exhibits a shift for one or two layers. Based on this observation, we propose selecting the layer $l$ with the following criteria:
$$
l = \text{argmax}_{l}\,\sum_{i=1}^n m\left( \mathbf{X}^i_{l\left\{-1\right\}}\right) + 1.
$$
\noindent{\bf Distance-aware MLE}
To mitigate the non-uniformity of density when applying MLE of the Poisson process, a common practice involves adjusting the rate $\lambda\left( t \right)$ of the Poisson process.~\citet{pmlr-v101-gomtsyan19a} suggest to replace the original rate $\lambda\left( t \right) = f V_m m t^{m - 1}$ as $\hat{\lambda}\left( t \right)  = f V_m m t^{m - 1} + t^m V_m \delta\left(t\right)$, where $\delta\left(R\right)$ is a correction function bounded by some geometric properties of the manifold of $\mathbf{X}^i$. For the exact form of $\delta\left(R\right)$, see~\citet{pmlr-v101-gomtsyan19a}. We will review the steps after the correction.

With the new rate, maximizing the log-likelihood in 
 Eq.~\ref{eq:likelihood} we will have:
\begin{equation}
\hat{m}\left(R, \mathbf{X}^i\right) = m\left(R, \mathbf{X}^i\right) \left(1 + \delta\left(R\right) \frac{R^2}{N\left(R, \mathbf{X}^i\right)}\right).
\label{eq:correction}
\end{equation}
 
Then, with Taylor expansion on the second term in Eq.~\ref{eq:correction},
we can calculate the correction with the following  polynomial regression:
\begin{equation}
\hat{m}\left(R, \mathbf{X}^i\right) = m\left(R, \mathbf{X}^i\right) + \sum_{j=1}^l \zeta_j R^j + \Theta \left(R^{l + 1}\right).
\label{eq:correction2}
\end{equation}

The new steps will be to estimate $\zeta_j$, and use the zero-order term as the estimated $\hat{m}\left(\mathbf{X}^i\right)$:
\begin{equation}
m\left(\mathbf{X}^i\right) = \hat{m}\left(\mathbf{X}^i\right) + \sum_{j=1}^l \zeta_j Q_T^j + \Theta \left(Q_T^{l + 1}\right).
\label{eq:correction2}
\end{equation}

In practice, solving the polynomial regression in Eq.~\ref{eq:correction2} can be conducted by minimizing the weighted least squared errors. We follow~\citet{pmlr-v101-gomtsyan19a} to use bootstrapping of $\mathcal{D}$: $\mathcal{D}_1, \mathcal{D}_2, \dots, \mathcal{D}_p$. We calculate the average of LIDs and distances, as well as the variance of LIDs.
$$
\bar{Q}_T = \frac{1}{p} \sum_{i=1}^{p}Q_{Tp},\, \bar{m}\left(\mathbf{X}^i\right) = \frac{1}{p} \sum_{i=1}^{p}m\left(\mathbf{X}^i\right)_{p},
$$
$$
\sigma(m(X^i)) = \frac{1}{p}\sum_{i=1}^p\left(m\left(\mathbf{X}^i\right)_{p} - \bar{m}\left(\mathbf{X}^i\right)\right)^2.
$$
Finally, the heteroskedastic weighted polynomial regression is minimized to find the final LID over different numbers of neighbors $T$:
$$
\text{min}\sum_{T=T_1}^{T_2}\frac{1}{\sigma(m(\mathbf{X}^i))} \left( \hat{m}\left(\mathbf{X}^i\right) - m(\mathbf{X}^i) - \sum_{j=1}^l \zeta_j Q_T^j\right)^2.
$$

We call the method with distance-aware MLE estimation as \textit{LID-GeoMLE}, and the vanilla MLE method \textit{LID-MLE}..

Clearly, for both LID-MLE and LID-GeoMLE, the estimated LID is a function of hyperparameters $T$, the number of neighbors, and $n$, the dataset size. While a small $T$ and $n$ provide an estimation of LIDs with large variance, a $T$ or $n$ that is too large will break the condition of local balls and neighbors. We elaborate in Section~\ref{section:rs} the effects of them in the performance.
\section{Experiments}
\begin{table*}[t]
\definecolor{lightgray}{gray}{0.9}
\centering
{
\begin{tabular}
{m{2.58cm}|m{1.5cm}<{\centering}m{1.5cm}<{\centering}m{1.55cm}<{\centering}|m{1.55cm}<{\centering}m{1.55cm}<{\centering}|m{1.55cm}<{\centering}|m{1.55cm}<{\centering}}
\toprule
Method & CoQA & TydiQA &\multicolumn{2}{c}{TriviaQA}  & \multicolumn{2}{c}{HotpotQA} & Averaged \\
 & 0-shot & 0-shot & 0-shot & 5-shot & 0-shot & 5-shot & 0-shot\\
\midrule
\rowcolor[gray]{0.95} 
  \multicolumn{8}{c}{Llama-2-7B} \cr 
Pred. Entropy & 0.715 & 0.590 & 0.697 & 0.768 & 0.650 & 0.669 & 0.663\\
LN-Pred. Entropy & 0.725 & 0.621 & 0.678 & 0.756 & 0.631 & 0.725 & 0.664\\
Semantic Entropy & 0.690 & 0.705 & 0.718 & 0.781 & 0.664 & 0.728 & 0.694\\
SAPLMA & 0.666 & 0.628 & 0.624 & 0.641 & 0.536 & 0.599 & 0.614\cr
P(True) & 0.638 & 0.608 & 0.471 & 0.651 & 0.444 & 0.593 & 0.540\\
\midrule
LID-MLE &  0.758 & 0.735 & 0.754 & 0.761& 0.701 & \bf 0.731 & 0.737\\
LID-GeoMLE & \bf 0.767  & \bf 0.738 & \bf 0.771 & \bf 0.791 & \bf 0.708 & 0.729 & \bf 0.746\\
\midrule
\rowcolor[gray]{0.95} 
  \multicolumn{8}{c}{Llama-2-13B} \cr 
Pred. Entropy & 0.745 & 0.630 & 0.751 & 0.752 & 0.738 & 0.765 & 0.716\\
LN-Pred. Entropy & 0.753 & 0.618 & 0.716 & 0.731 & 0.724 & 0.769 & 0.702\\
Semantic Entropy & 0.758  & 0.740 & 0.736 & 0.786 & 0.708 & \bf 0.781 & 0.736\\
SAPLMA & 0.645 & 0.597 & 0.651 & 0.699 & 0.578 & 0.621 & 0.618\cr
P(True) &  0.649 & 0.624 & 0.511 & 0.662 &  0.518 & 0.581 & 0.576\\
\midrule
LID-MLE & 0.763 & 0.745& 0.748 & 0.777 & 0.747 & 0.758 & 0.751\\
LID-GeoMLE & \bf 0.772& \bf 0.759 & \bf 0.775  & \bf 0.793 & \bf 0.749 & 0.769 & \bf 0.764\\
\bottomrule
\end{tabular}
}
\caption{Main results of predicting output correctness for four generative QA tasks. We compare our LID methods: LID-MLE, LID-GeoMLE with entropy-based and verbalized uncertainty estimation methods and a trained classifier. We show the results with Llama-2 7B and 13B. Results demonstrate the superior performance of our LID methods. The best scores are \textbf{bold}.}
\label{tab:main}
\vspace{-3pt}
\end{table*}

In this section, we empirically demonstrate the effectiveness of using LID to predict the truthfulness of model outputs, outperforming uncertainty-based methods and classifiers trained to predict truthfulness.

\noindent{\bf Datasets \& Models} We consider four generative QA tasks:
TriviaQA~\citep{joshi2017triviaqa}, CoQA~\citep{reddy2019coqa}, HotpotQA~\citep{yang2018hotpotqa}, and TydiQA-GP (English)~\citep{clark2020tydi}.
These datasets cover different formats of QA including open-book (CoQA), closed-book (TriviaQA, HotpotQA), and reading comprehension (TydiQA-GP) and several capacities of LLMs. For each of the datasets, we generate outputs for 2,000 samples from the validation sets and test the methods with those samples.

We evaluate with the decoder-only Transformer-based Llama-2~\citep{touvron2023llama2}, 7B and 13B, which are cutting-edge public foundation models whose internal representations are accessible. In our preliminary experiments, we have also tested on Llama, OPT~\citep{zhang2022opt} and have similar observations as Llama-2. Following the inference convention as in~\citet{touvron2023llama2}, we conduct both zero-shot and few-shot inference. See inference examples with our prompt format in Appendix~\ref{app:dataset}.

\noindent{\bf Methods \& Baselines} For LID-MLE and LID-GeoMLE, we use 500 nearest neighbors when estimating LIDs for all datasets. We compare our methods with entropy-based uncertainty, verbalized uncertainty, and trained truthfulness classifiers on representations. 

More specifically, for entropy-based ones, we consider predictive entropy (\textit{Pred. Entropy}), length-normalized predictive entropy (\textit{LN-Pred. Entropy})~\citep{malinin2020uncertainty}, and semantic entropy (\textit{Semantic Entropy})~\citep{kuhn2022semantic}. Since precisely calculating the entropy is intractable over the infinite output space of generative QA tasks, we use Monte Carlo estimation of entropy on sampled outputs~\citet{malinin2020uncertainty}. Formally, 
$
\text{Pred. Entropy}\,=\,\frac{1}{N} \sum_{i=1}^{N} \text{log} p\left(y_i | x \right)
$, where $y_i,\,i=1 \dots N$ is the N sampled outputs. LN-Pred. Entropy simply replaces the log-likelihood with the length-normalized log-likelihood:
$
\text{LN-Pred. Entropy}\,=\,\frac{1}{N} \sum_{i=1}^{N} \frac{1}{|y_i|} log p\left(y_i | x \right)
$. Semantic Entropy groups outputs that are semantically equivalent to each other and calculate the entropy among groups: $
\text{Semantic Entropy}\,=\,\frac{1}{|C|} \sum_{i=1}^{|C|} log p\left(C_i | x \right)
$, where $C_i$ is the summed likelihood of outputs in the i-th group. All entropy-based methods are sensitive to the choice of temperature in decoding and the number of sampled outputs. We follow~\citet{kuhn2022semantic}, and set the temperature to be 0.5 and the number of generated samples to be 10.

For verbalized uncertainty, we use $P\left(\text{True}\right)$~\citep{kadavath2022language}, which asks the model itself if its answer is correct. Then, take the probability for the model outputting the token `True' as the truthfulness score.

For trained classifiers, we implement \textit{SAPLMA}~\citep{azaria2023internal} that train a multi-layer classifier on each dataset with 3,000 examples (1,500 truthful and 1,500 untruthful) to predict a binary truthfulness label for each example. The training setup follows~\citet{azaria2023internal}.

\noindent{\bf Evaluation Setup}
We use the area under the receiver operator
characteristic curve (AUROC) to evaluate the effectiveness of all baselines and our proposed LID method. The truthfulness prediction task is viewed as a binary classification task, and AUROC measures the performance under varying thresholds. The indicator function of truthfulness is given by $s\left(y_i, \hat{y}_i\right)\,=\,\mathbb{I}\left( \text{RougeL}\left(y_i, \hat{y}_i\right) \geq 0.5 \right)$, following~\citep{kuhn2022semantic}, where $\text{RougeL}$~\citep{lin2004rouge} score is a substring matching measurement commonly used to evaluate generative QA tasks. We show that the results is robust across different indicator function in Appendix~\ref{app:indicator}.

For TriviaQA and HotpotQA, we evaluate with both zero-shot and 5-shot in-context inference. We evaluate CoQA and TydiQA-GP with zero-shot as the context is long.

\subsection{Sanity Check}
To gain insights about the reliability of MLE-based estimators, we apply MLE-based methods on synthetic data with known ground truth dimensions, showing that they can approximately give the correct estimation. Additionally, we compare the intrinsic dimension obtained by MLE-based methods and other popular methods: KNN~\citep{Costa2005EstimatingLI} and TwoNN~\citep{Facco2017EstimatingTI}.

For synthetic data, we simulate two popular manifolds, namely a sphere and a norm, both having an original dimension of 4,096 and intrinsic dimensions of 10 and 20 (Table~\ref{table:sanity}). We find that GeoMLE in general produces the most accurate estimation on those datasets, although MLE shows more negative bias. For real datasets, TwoNN and MLE-based methods give approximately similar estimations. Overall, MLE-based methods produce reasonable estimates in both scenarios. It is safe to assume that MLE offers a practical approximation for applications, especially when the true value of intrinsic dimension is not directly relevant to the application but rather comparisons of the values. Furthermore, note that MLE inherently estimates `local' intrinsic dimension while other methods are `global' estimators. Based on the points above, we adopt MLE-based methods in our study.

\begin{table}[t]
\renewcommand{\arraystretch}{1.0}
\begin{tabular}[t]{p{1.8cm} | p{0.4cm} p{0.8cm} p{0.7cm}p{0.7cm}p{1.15cm}}
\toprule 
\bf Dataset & m & TwoNN & KNN & MLE & GeoMLE \cr
\toprule 
Sphere	&10	&8.78	&4	&8.63	&8.65\cr
Sphere noise	&10	&13.97	&4	&11.45	&9.64\cr
Norm	&20	&17.54	&9	&15.54	&20.33\cr
Norm noise	&20	&17.81	&2	&15.72	&22.36\cr
\midrule
\multirow{2}{2em}{TriviaQA} & T & 10.10 & 5	& 9.37 & 8.68\cr	
& F & 15.45 & 10 & 12.43 & 11.97\cr
\multirow{2}{2em}{CoQA} & T & 16.21 & - & 14.50 & 13.98\cr	
& F & 17.79 & - & 16.62 & 16.10 \cr
\bottomrule
\end{tabular}
\caption{Sanity Check on intrinsic dimension estimate. We compare MLE with other famous estimates on synthetic data and our QA datasets. All with 1000 samples. For synthetic data, the column `m' represents the ground-truth intrinsic dimensions. For simulated data, we report results on truthful (T) and untruthful (F) samples. KNN cannot provide reasonable estimates on CoQA.}
\label{table:sanity}
\end{table}

\subsection{Main Results}
The results are summarized in Table \ref{tab:main}. See Appendix~\ref{app:hist} for the corresponding frequency histograms. Overall, we observe that LID-GeoMLE outperforms entropy-based methods by 0.05 points for 7B and 0.03 points for 13B on AUROC, while verbalized uncertainty are not comparable to the above methods. Notably, the performance of uncertainty-based methods is contingent on whether we present in-context examples to the LLMs while LID methods are more stable regarding this. For example, on TriviaQA with 0-shot, LID methods outperform the best entropy-based uncertainty estimation by 8\%. The improvements comparing LID-MLE and LID-GeoMLE suggest that a more sophisticated LID estimation method would bring additional benefits to the truthfulness prediction performance.

Moreover, notice that another representation-level method, SAPLMA, fails to match the performance of LID-based methods. This implies that LID-based methods remain powerful when truthful directions are hard to obtain. We show in Appendix~\ref{app:tsne} that dimension reduction methods like t-SNE also fail to find such directions. The findings in~\citep{marks2023geometry} might not hold in a practical scenario.

\definecolor{LightCyan}{rgb}{0.88,1,1}
\definecolor{lightgray}{gray}{0.9}
\begin{table}[t]
\renewcommand{\arraystretch}{1.05}
\scalebox{1}{
\begin{tabular}[t]{p{1.4cm} | p{1.1cm} p{1.2cm} p{1.1cm}p{1.1cm}}
\toprule 
\bf & TriviaQA & HotpotQA & TydiQA & CoQA \cr
\toprule 
TriviaQA &	\cellcolor{lightgray}0.748&	0.729&	0.734&	0.745\cr

HotpotQA &	0.724&	\cellcolor{lightgray}0.747 &	0.726&	0.724\cr
TydiQA &	0.765 $\uparrow$ &	0.744 &	\cellcolor{lightgray}0.745 &	0.736\cr
CoQA &	0.747&	0.753 $\uparrow$&	0.751 $\uparrow$&	\cellcolor{lightgray}0.763 \cr
\bottomrule
\end{tabular}}
\caption{Robustness to cross-task neighbors. Performance when the neighbors are from different datasets. The left column is the neighbors' dataset while the top column shows the tested dataset. The performance in general decreases slightly but is still effective.}
\label{table:crosstask}
\end{table}
\begin{figure}[h]
    \centering
    \includegraphics[scale=0.35]{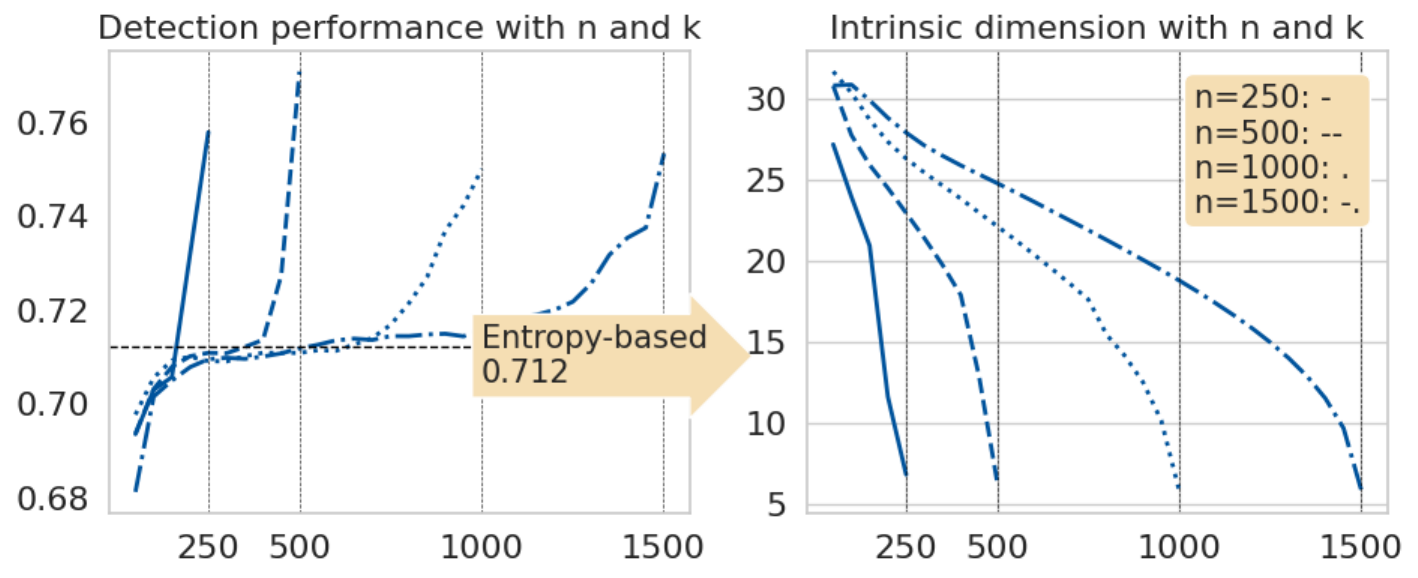}
    \caption{Robustness to $n$ and $T$. Performance and intrinsic dimension as a function of the number of neighbors and total reference points. Both plots use the number of neighbors as X-axis. Different line styles indicate different numbers of reference points.}
    \label{fig:function}
\end{figure}

\begin{figure*}[h]
    \centering
    \includegraphics[scale=0.47]{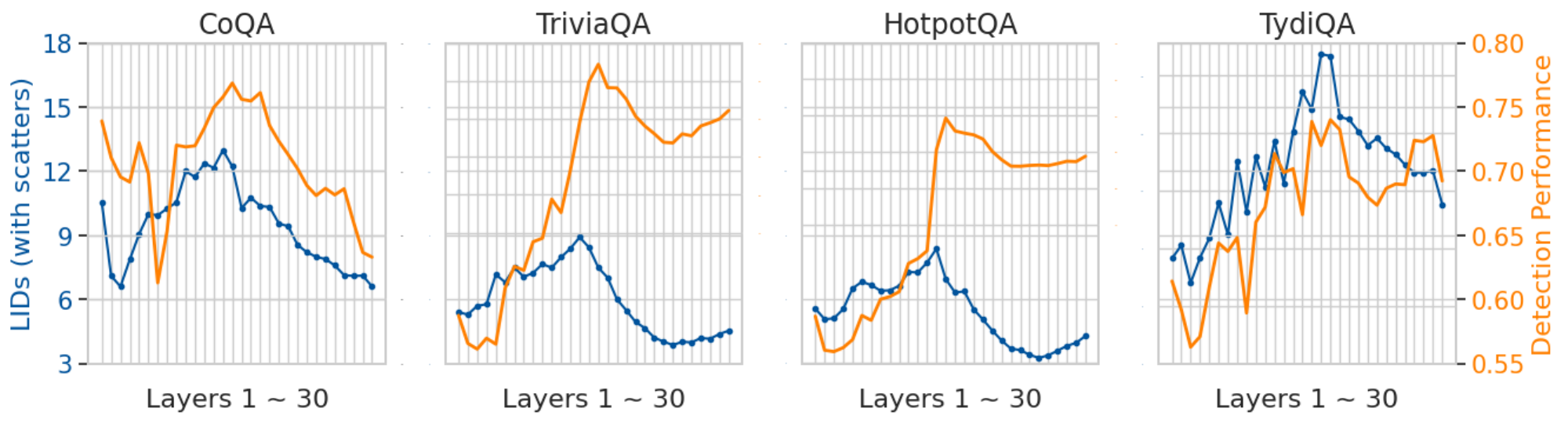}
    \caption{Plots for the aggregated LID values across model layers on the four QA datasets. The X-axis is the layer id, which is layer 1 to layer 30 for Llama-2-7B. The left Y-axis is the aggregated LID values and the right Y-axis is the detection performance (AUROC) values. The detection performance curve is in orange and the LID curve is in blue with markers. We show that there is a hunchback shape in the LID values across layers. The LID values closely correlate with the performance of detection and exhibit a `shift behind' phenomenon.}
    \label{fig:delaycurve}
\end{figure*}
\subsection{Robustness Study}
\label{section:rs}

We study the robustness of our proposed LID technique. 

\noindent{\bf Robustness to hyperparameters $n$ and $T$.} We investigate how the performance and LIDs vary as a function of different numbers of dataset size $n$ and neighbors $T$.

As illustrated in Figure~\ref{fig:function}, the performance on AUROC increases as we consider more neighbors, reaching its maximum as the number of neighbors used approaches the total dataset size. The performance of using LID method is comparable to or better than the best entropy-based uncertainty estimation methods even with around 200 neighbors.

The intrinsic dimension shows a decreasing trend when we consider more neighbors, moving within the range of 6 to 30. The intrinsic dimension value will be lower with the same number of neighbors but with a larger dataset size. Again, MLE methods provide just approximations to the actual intrinsic dimension, which may contain bias but is of no direct concern in this application of detecting hallucinations.

\noindent{\bf Robustness to cross-task reference}
To understand how generalizable the LID feature is and whether we can use LIDs in the wild, we conduct experiments where the neighbors come from a different dataset. Results are shown in Table~\ref{table:crosstask}. We find that there is only a small decrease in performance in general, demonstrating the features used to estimate intrinsic dimension are generalizable to out-of-domain tasks. Also, notice that we observe some performance boosts when the reference samples come from a dataset that models have better performance on. For example, TydiQA reference samples improve the detection on HotpotQA. We leave further investigation to this observation to future work.


\section{Analysis}

In this section, we conduct a series of analyses on the characteristics of intrinsic dimensions of LLM representations. In the first two parts, we study how the intrinsic dimensions of model representations change across layers and during the autoregressive language modeling process. In the last part, we investigate how the effects of instruction tuning on intrinsic dimensions. Our study reveals that intrinsic dimension is indeed an insightful tool to understand LLMs.

\subsection{The aggregated LIDs exhibit a hunchback shape in intermediate layers}

To study the characteristics of the intrinsic dimensions insider the model layers, we aggregate individual LIDs and present the trends of the aggregated LIDs across different model layers. As depicted in Figure ~\ref{fig:delaycurve}, across the four datasets, the intrinsic dimensions exhibit a \textit{hunchback} shape, akin to the observation in the vision domain~\citep{ansuini2019intrinsic}: the averaged LID values initially increase from the bottom to middle layers, and then decrease from the middle to the top layers. Note that intrinsic dimensions represent how many dimensions are needed to encode the information without significant loss. The observed hunchback phenomenon suggests that LLMs gradually capture the information in the context in the first few layers, and then condense them in the last layers to map to the vocabulary.

Furthermore, we observe a close relation between the intrinsic dimension values and the performance of predicting individual truthfulness. The two curves: the LID curve with dots (blue) and the detection performance curve (orange) show similar trends. Nevertheless, there is a `shifting behind' effect: the variants in LID values are reflected one or two layers later in the prediction of truthfulness. We hypothesize that once the model encodes sufficient information, as indicated by the absolute LID values, additional transformations in later blocks are required to convert these encoded features into indicators of truthfulness. Empirical verification and further investigation of this phenomenon could be explored in future work.

\subsection{The intrinsic dimensions are consistently lower for human answers at different positions}

\begin{figure}[t]
    \centering
    \includegraphics[scale=0.41]{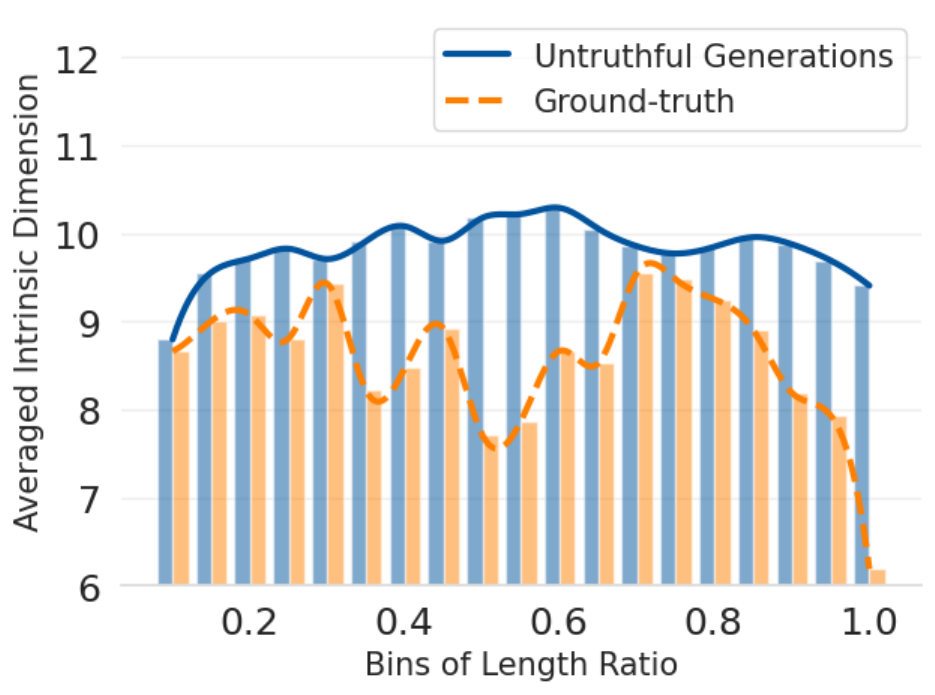}
    \caption{Bars for intrinsic dimensions of ground-truth answers (orange dashed line) and untruthful model generations (blue line) as the language modeling proceeds. The X-axis represents buckets of different ratios of the total lengths.}
    \label{fig:distcurve}
\end{figure}

Next, to investigate how intrinsic dimensions vary when modeling truthful and untruthful outputs, we compare the LIDs of untruthful answers with the LIDs of their corresponding ground-truth answers at each position, using another set of complete correct answers as reference points. We use the TriviaQA dataset as an example. Figure~\ref{fig:distcurve} illustrates the aggregated results. We observe that the intrinsic dimensions of ground-truth are consistently lower than model generations at different positions. For ground-truth answers, a sharp decrease in the intrinsic dimensions happens when approaching the end of generations while this phenomenon does not hold for incorrect generations. This explains why selecting the last token gives the best performance when working with representations~\citep{zou2023representation}.

To conduct a controlled study, we construct a few examples where we explicitly mix human and model answers: we prompt the model with the first half of the correct answers but ask the model to generate the rest. As in Table~\ref{table:mix}, the mixed answers usually have higher LIDs compared to both human and model answers, even when the model answer is incorrect. This supports our hypothesis that incorrect answers mix more manifolds and thus have higher LIDs. More examples are displayed in Appendix~\ref{app:mix}.
\begin{table}[h]
\renewcommand{\arraystretch}{1.0}
\begin{center}
\small
\begin{tabular}{p{7.55cm}}
\toprule
\texttt{Answer these questions: Q: Who played the title roll in the Flint films? A: }\cr
\midrule
- \textbf{Model output: } \texttt{John Cusack} $\left[\text{7.54, 8.61, 13.34, \bf 5.05}\right]$\cr
- \textbf{Ground-truth: } \texttt{James Coburn} $\left[\text{7.88, 9.96, \bf 4.79}\right]$ \cr
- \textbf{Mixed output: } \texttt{James {\color{blue}Garner}}  $\left[\text{7.87,  8.72,  4.58 , \bf 6.98}\right]$ \cr

\bottomrule\hline
\end{tabular}
\vspace{-6pt}
\end{center}
\caption{Examples of mixing distributions increases LIDs. The {\color{blue}Blue} part is the model continuation for the ground-truth. The numbers in the list represent LID value for each position. }
\label{table:mix}
\end{table}

\subsection{The intrinsic dimensions increase while instruction tuning and correlate with model performance}
\begin{figure}[t]
    \centering
    \includegraphics[scale=0.39]{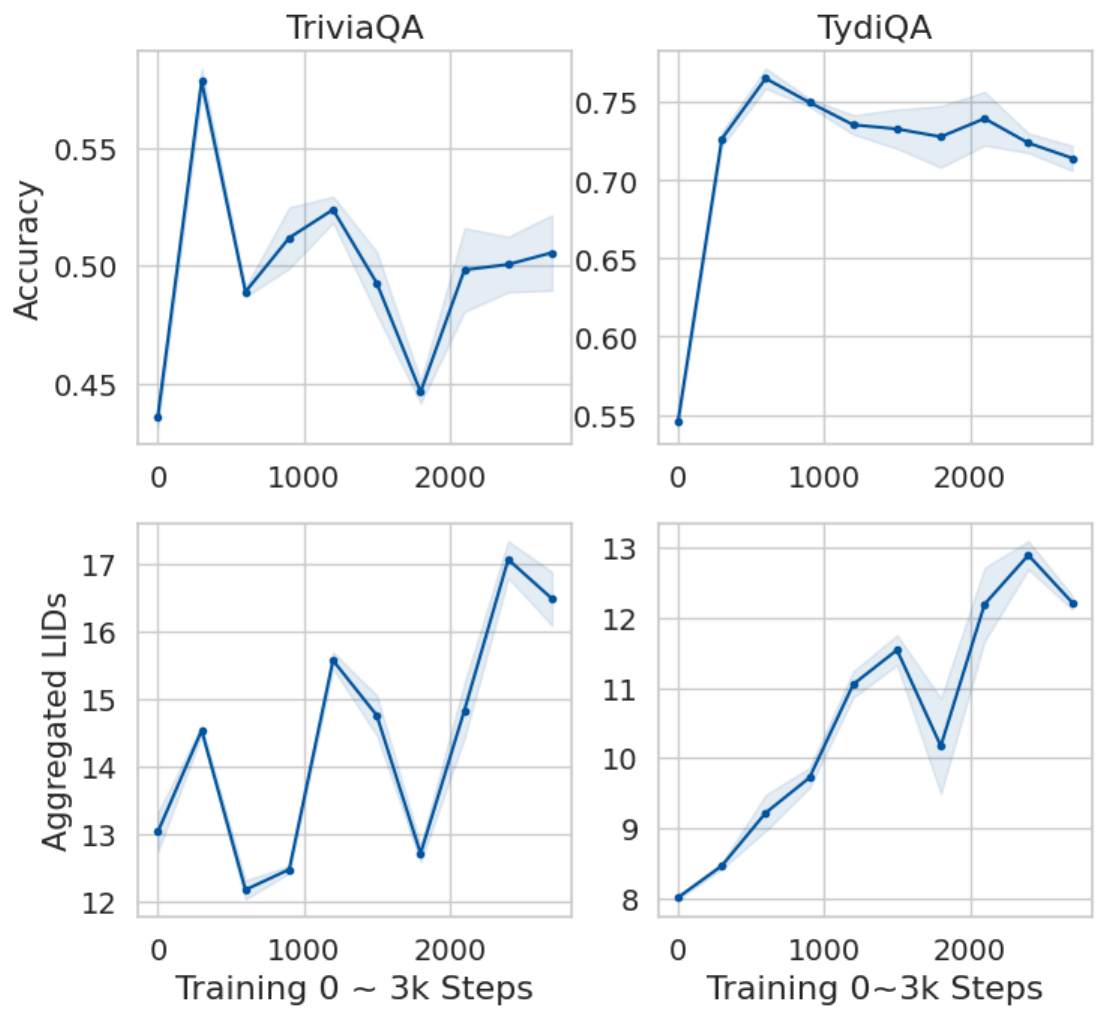}
    \caption{Plots for the accuracy and intrinsic dimension on TriviaQA and TydiQA during instruction tuning. The X-axis is the training steps. We train 3,000 steps in total and show checkpoints every 300 steps. The Y-axis is the performance for the top two figures and the aggregated LID values for the bottom two figures.}
    \label{fig:ITcurve}
\end{figure}
Finally, we study how instruction tuning affects LLMs' intrinsic dimensions. Instruction tuning adapts a pre-trained LLM to solve diverse tasks by training LLMs to follow declarative human instructions~\citep{supernaturalinstructions, alpaca}. We follow this paradigm and investigate the representation LIDs.

\noindent{\bf Experimental Setup}
We use the \textsc{Super-NI}~\citep{supernaturalinstructions} for training, which contains 756 training tasks and 200 examples for each training task. We fine-tune a Llama-2-7B model for 3,000 steps, roughly 3 epochs, on \textsc{Super-NI}'s training set. We track every 300 steps during the tuning process and evaluate those checkpoints for their accuracy on TydiQA and TriviaQA, as well as the intrinsic dimension footprints. For both TydiQA and TriviaQA, we randomly sample 1,000 test examples and repeat the experiments three times. We use $T$=500 nearest neighbors for estimating the LIDs on TriviaQA and TydiQA.

\noindent{\bf The intrinsic dimension grows while training for a longer time}
As illustrated in Figure~\ref{fig:ITcurve}, instruction tuning brings a performance boost for both TriviaQA and TydiQA, while the boost is more significant for TydiQA. We also find that the aggregated LID values show an increasing trend along with the training process, although there are more fluctuations for TriviaQA than TydiQA. During instruction tuning, models are tuned on diverse tasks and composed distributions. The increase in intrinsic dimensions of LLM representations possibly implies that they become richer.

\noindent{\bf The aggregated LID values correctly predict fluctuations in generalization ability during training}
We observe that the performance on both TriviaQA and TydiQA reaches a local minimum at some intermediate checkpoints during instruction tuning, while the intrinsic dimensions decrease correspondingly. For example, at step 600 and step 1800, the performance for TriviaQA fluctuates and reaches the local minimum. This is reflected in the curve of LID values, where at step 600 and step 1800, the LID values are the lowest locally. Similar observations exist for TydiQA at step 1800. This suggests that one may use the intrinsic dimension as a signal to select model checkpoints.
\section{Conclusion}
In this paper, we proposed to use LIDs to characterize and predict the correctness of LLM outputs, which achieved better performance compared to prior methods. We showed several empirical observations about model intrinsic dimensions, including the variants of them with model layers, autoregressive language modeling, and the effects of instruction tuning. This opened up a new direction to consider quantifying model truthfulness for future work.

\section{Impact Statement}
This paper discusses an important step for the safe deployment of LLMs. LLMs have demonstrated their remarkable capabilities, but human trust on LLMs are still limited because of their tendency to generate hallucinations secretly. We propose to quantify and characterize model hallucinations through intrinsic dimensions of model intermediate representations. On the one hand, this method itself helps people abstain from trusting incorrect generations. On the other hand, it can serve as the backbone for some hallucination mitigation methods.

We believe the method can be extended to other settings besides detecting hallucinations. For example, it might be possible to leverage characteristics in geometry or intrinsic dimensions to detect harmful prompts or model generations, It might also be possible to detect toxicity or adversarial data with intrinsic dimensions.

This paper mainly discusses LLM generations in English. Results may be biased towards the English-speaking population. However, adapting this LID method to other languages does not require much additional efforts. Moreover, the paper focuses primarily on question-answering scenarios. We encourage future work to implement this method on diverse tasks and languages, as well as cross-lingual settings.

We will release the data, code, as well as representation and model checkpoints to facilitate reproduction of the results in this paper.

\section*{Acknowledgement}
We thank UCLA-NLP and UCLA-PlusLab members for their invaluable feedback while preparing this draft. We thank Po-Nien Kung for providing the codebase of instruction tuning. This project is supported by a sponsored research award by Cisco Research.

\nocite{langley00}

\bibliography{example_paper}
\bibliographystyle{icml2024}


\clearpage
\appendix
\onecolumn
\section{Dataset and Inference Details}
\label{app:dataset}
For datasets without context (HotpotQA and TriviaQA), we use the following textual input as prompts:
$$
\textit{Answer these questions: \textbackslash n Q: [question] \textbackslash n A: }
$$
For datasets with context (TydiQA-GP and CoQA), we have the following template for prompts:
$$
\textit{Answer these questions based on the context:\textbackslash n 
Context: [a passage or a paragraph] \textbackslash n
Question: [question]
Answer: }
$$
\begin{table}[h]
\renewcommand{\arraystretch}{1}
\begin{center}
\small
\begin{tabular}{p{15.85cm}}
\toprule
TriviaQA \cr
\midrule

- Answer the following questions: \textbackslash n
Q: The Great Barrier Reef is located off the coast of which Australian state? \textbackslash n
A: \cr
- Answer the following questions: \textbackslash n
Q: Whose 2006 album ""Back to Black"" had 6 Grammy Award nominations and five wins, tying the record for the most wins by a female artist in a single night, and made her the first British singer to win 5 Grammys?\textbackslash n
A: \cr
- Answer the following questions: \textbackslash n
Q: Who directed The Cable Guy? \textbackslash n
A: \cr
\midrule
HotpotQA \cr
\midrule
- Answer these questions: \textbackslash n
Q: Which film was created first, Tex or Miracle? \textbackslash n
A: \cr
- Answer these questions: \textbackslash n
Q: When was the company, of which Ernest Walter Hives was one time chairman, founded ? \textbackslash n
A: \cr
\midrule
CoQA \cr
\midrule
- Answer these questions based on the context: \textbackslash n
Context: Annie was helping her little brother Max pick flowers from the garden. They wanted to put the flowers in a jar to put on the kitchen table. Mother's Day was the next day and their mother loved fresh flowers. After they picked flowers and put them in a jar, Max asked Annie if they could have a snack. Annie took Max into the kitchen and got out an apple to slice up. They sat down at the table looking at the flowers and ate their apple slices. There was a window in the kitchen that let in sunlight. ""Hey!"" Max said, pointing at one of the roses in the jar. ""There's something moving on that rose."" Annie looked more closely at the flowers. ""It's a ladybug,"" she said. ""We need to take it back outside."" Suddenly the ladybug began flying around the kitchen. Max jumped up and ran around trying to catch it. At last he clapped his hands around it. ""Careful!"" said Annie. Max walked outside and let the ladybug go. \textbackslash n
Question: Was Max an older brother? \textbackslash n
Answer: \cr
\midrule
TydiQA-GP \cr
\midrule
- Answer the following question based on the information in the given passage: \textbackslash n 
Passage: The Kingdom of Aksum (also known as the Kingdom of Axum, or the Aksumite Empire) was an ancient kingdom located in what is now Tigray Region (northern Ethiopia) and Eritrea.[2][3]. Axumite Emperors were powerful sovereigns, styling themselves King of kings, king of Aksum, Himyar, Raydan, Saba, Salhen, Tsiyamo, Beja and of Kush.[4] Ruled by the Aksumites, it existed from approximately 100 AD to 940 AD. The polity was centered in the city of Axum and grew from the proto-Aksumite Iron Age period around the 4th century BC to achieve prominence by the 1st century AD.  Aksum became a major player on the commercial route between the Roman Empire and Ancient India. The Aksumite rulers facilitated trade by minting their own Aksumite currency, with the state establishing its hegemony over the declining Kingdom of Kush. It also regularly entered the politics of the kingdoms on the Arabian Peninsula and eventually extended its rule over the region with the conquest of the Himyarite Kingdom. The Manichaei prophet Mani (died 274 AD) regarded Axum as one of the four great powers of his time, the others being Persia, Rome, and China.[2][5] \textbackslash n
Question: When did the Kingdom of Aksum end? \textbackslash n
Answer: \cr
\bottomrule
\end{tabular}
\vspace{-6pt}
\end{center}
\caption{Examples of datasets.}
\label{table:examples}
\end{table}
Table~\ref{table:examples} shows some examples from those datasets with our inference format. 
\definecolor{LightCyan}{rgb}{0.88,1,1}
\definecolor{lightgray}{gray}{0.9}

\begin{table}[h]
\renewcommand{\arraystretch}{1.05}
\begin{center}
\small
\scalebox{1}{
\begin{tabular}[t]{p{2.4cm} | p{1.1cm} p{1.2cm} p{1.1cm}p{1.1cm}}
\toprule 
\bf & TriviaQA & HotpotQA & TydiQA & CoQA \cr
\toprule 
Llama-2-7B &	0.66&	0.26&	0.44&	0.57\cr

Llama-2-13B &	0.72&	0.33 &	0.50&	0.62\cr

\bottomrule
\end{tabular}}
\end{center}
\caption{Accuracy of Llama-2-7B and Llama-2-13B on the four datasets with $s\left(y_i, \hat{y}_i\right)\,=\,\mathbb{I}\left( \text{RougeL}\left(y_i, \hat{y}_i\right) \geq 0.5 \right)$}
\label{table:performance}
\end{table}
The performance with LLama-2 is shown in Table~\ref{table:performance}, which roughly matches the publicly reported performance.
\clearpage

\section{Different Indicator Functions}
\label{app:indicator}
We evaluate the sensitivity of our methods towards different indicator functions. We consider two new indicator functions: 

1)  $s\left(y_i, \hat{y}_i\right)\,=\,\mathbb{I}\left( \text{RougeL}\left(y_i, \hat{y}_i\right) \geq 0.3 \right)$, i.e., changing the thresholds in the Rouge-L metric from 0.5 to 0.3;

2) $s\left(y_i, \hat{y}_i\right)\,=\,\mathbb{I}\left( \text{NLI}\left(y_i, \hat{y}_i\right) == \text{entailment} \right)$. We use a natural language inference (NLI) model to judge the semantic similarity of two answers as the indicator function. The model is tuned for judging whether a premise semantically entails a hypothesis, where the ground-truth answer is used as the premise and the model-generated answer is used as the hypothesis. The NLI model is based on T5-XXL from~\citet{honovich-etal-2022-true-evaluating}.

We evaluate with TriviaQA and CoQA on Llama-2-7B. The only difference with our main results is that the indicator functions are changed to the above two functions. Results are shown in Table~\ref{tab:other}. We show that despite of small performance variants, LID-MLE and LID-GeoMLE outperform the best uncertainty-based methods across different indicator functions.
\begin{table*}[h]
\definecolor{lightgray}{gray}{0.9}
\centering
{
\begin{tabular}
{m{3.58cm}|m{2.0cm}<{\centering}m{2.0cm}<{\centering}m{2.0cm}<{\centering}}
\toprule
Method & CoQA & TriviaQA 0-shot & TriviaQA 5-shot \\
\midrule
\rowcolor[gray]{0.95} 
  \multicolumn{4}{c}{Rouge-L $\geq$ 0.3} \cr 
\midrule
Semantic Entropy &  0.683 & 0.726 & 0.791\cr
SAR~\citep{duan2023shifting} & 0.694 & 0.731 & 0.795 \cr
LID-MLE & 0.744 & 0.761 & 0.783\cr
LID-GeoMLE & \bf 0.756 & \bf 0.784 & \bf 0.801\cr
\midrule
\rowcolor[gray]{0.95} 
  \multicolumn{4}{c}{NLI Entailment} \cr 
\midrule
Semantic Entropy & 0.666 & 0.715 & 0.779\cr
SAR~\citep{duan2023shifting} & 0.671 & 0.758 & 0.783\cr
LID-MLE & 0.723 & 0.762 & 0.783\cr
LID-GeoMLE & \bf 0.735 & \bf 0.781 & \bf 0.790\cr
\bottomrule
\end{tabular}
}
\caption{Detect incorrect answers for Llama-2-7B on TriviaQA and CoQA. We use two other indicator functions based on Rouge-L threshold of 0.3 and NLI entailment. Results demonstrate that LID methods is robust to different indicator functions.}
\label{tab:other}
\vspace{-3pt}
\end{table*}

\clearpage

\section{Frequency Histogram}
\label{app:hist}
\begin{figure}[h]
    \centering
    \includegraphics[scale=0.43]{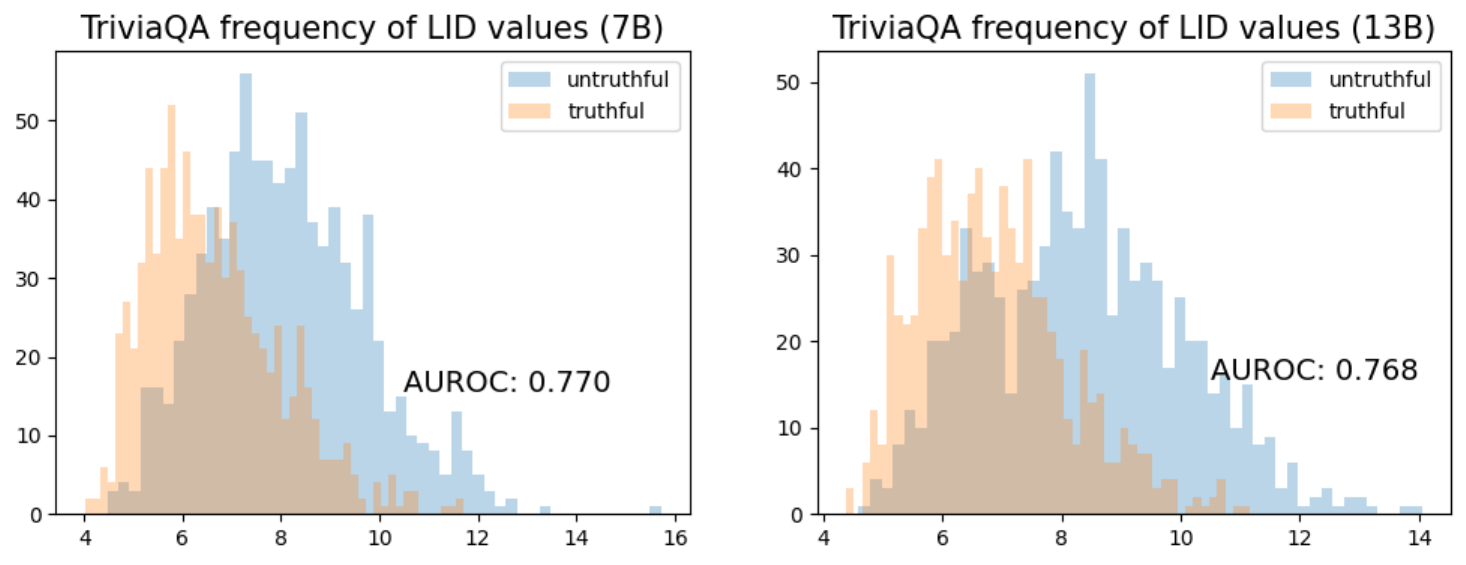}
    \caption{Frequencies of the LID values for truthful and untruthful data on TriviaQA, with Llama-2-7B (left) and Llama-2-13B (right). X-axis is the LID values while Y-axis is the number of occurs.}
    \label{fig:histgram}
\end{figure}

\begin{figure}[h]
    \centering
    \includegraphics[scale=0.43]{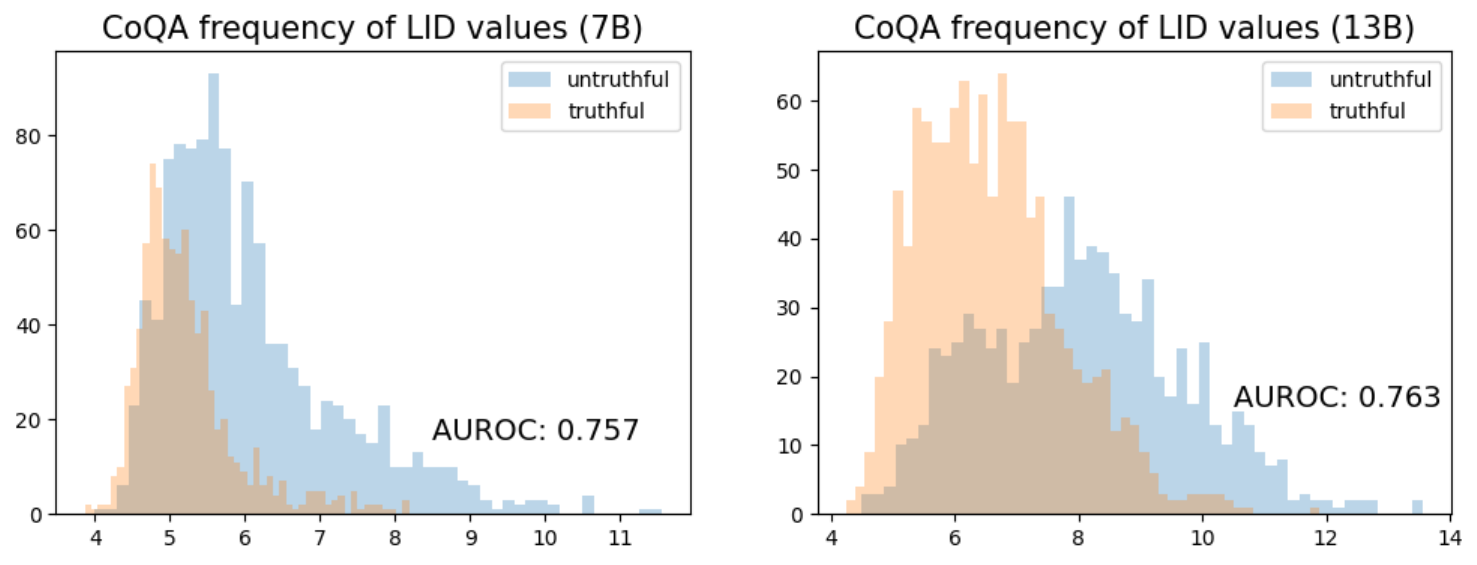}
    \caption{Frequencies of the LID values for truthful and untruthful data on CoQA, with Llama-2-7B (left) and Llama-2-13B (right). X-axis is the LID values while Y-axis is the number of occurs.}
    \label{fig:histgram}
\end{figure}

\begin{figure}[h]
    \centering
    \includegraphics[scale=0.43]{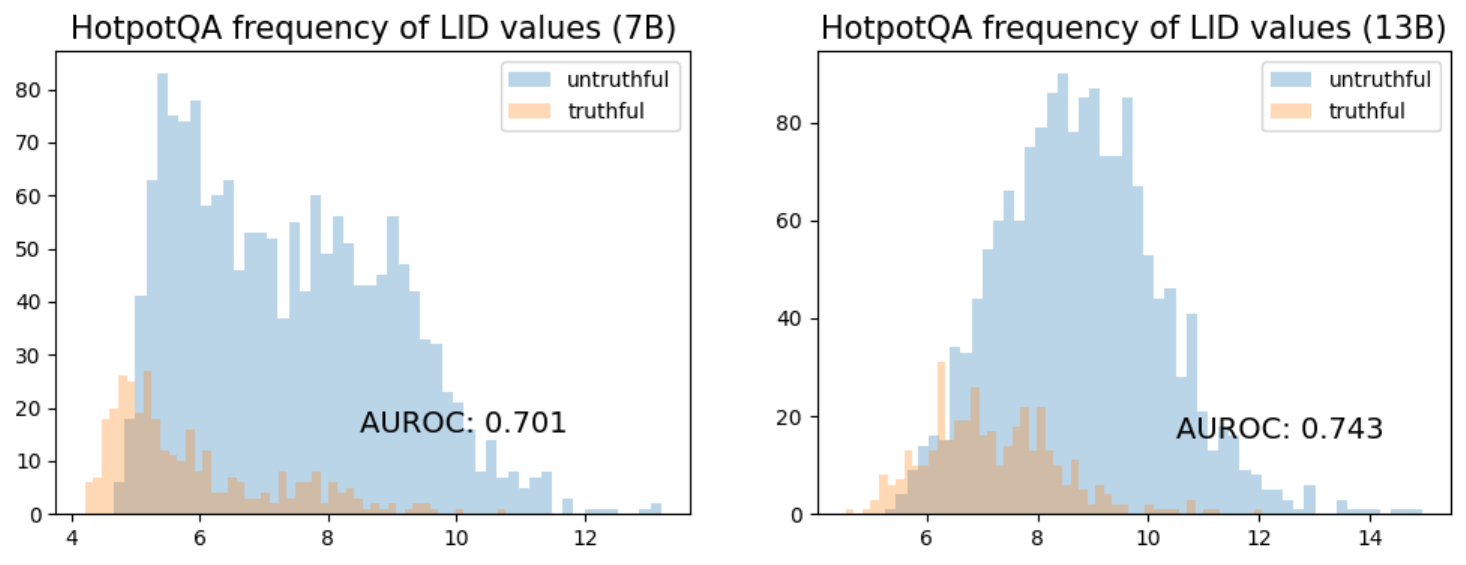}
    \caption{Frequencies of the LID values for truthful and untruthful data on HotpotQA, with Llama-2-7B (left) and Llama-2-13B (right). X-axis is the LID values while Y-axis is the number of occurs.}
    \label{fig:histgram}
\end{figure}

\clearpage
\section{Visualization of truthful and untruthful answers with t-SNE}
\label{app:tsne}
We show the t-SNE scatters that reduce the dimensions of original representations into a two-dimensional space. On CoQA and TydiQA, there are vague clusters of truthful and untruthful generations. But on TriviaQA and HotpotQA, there is no clear cluster. Overall, the visualization shows that dimension reduction methods like t-SNE fail to distinguish truthful answers from untruthful answers.
\begin{figure}[!htb]
    \centering
    \begin{minipage}{.5\textwidth}
        \centering
        \includegraphics[width=0.9\linewidth, height=0.3\textheight]{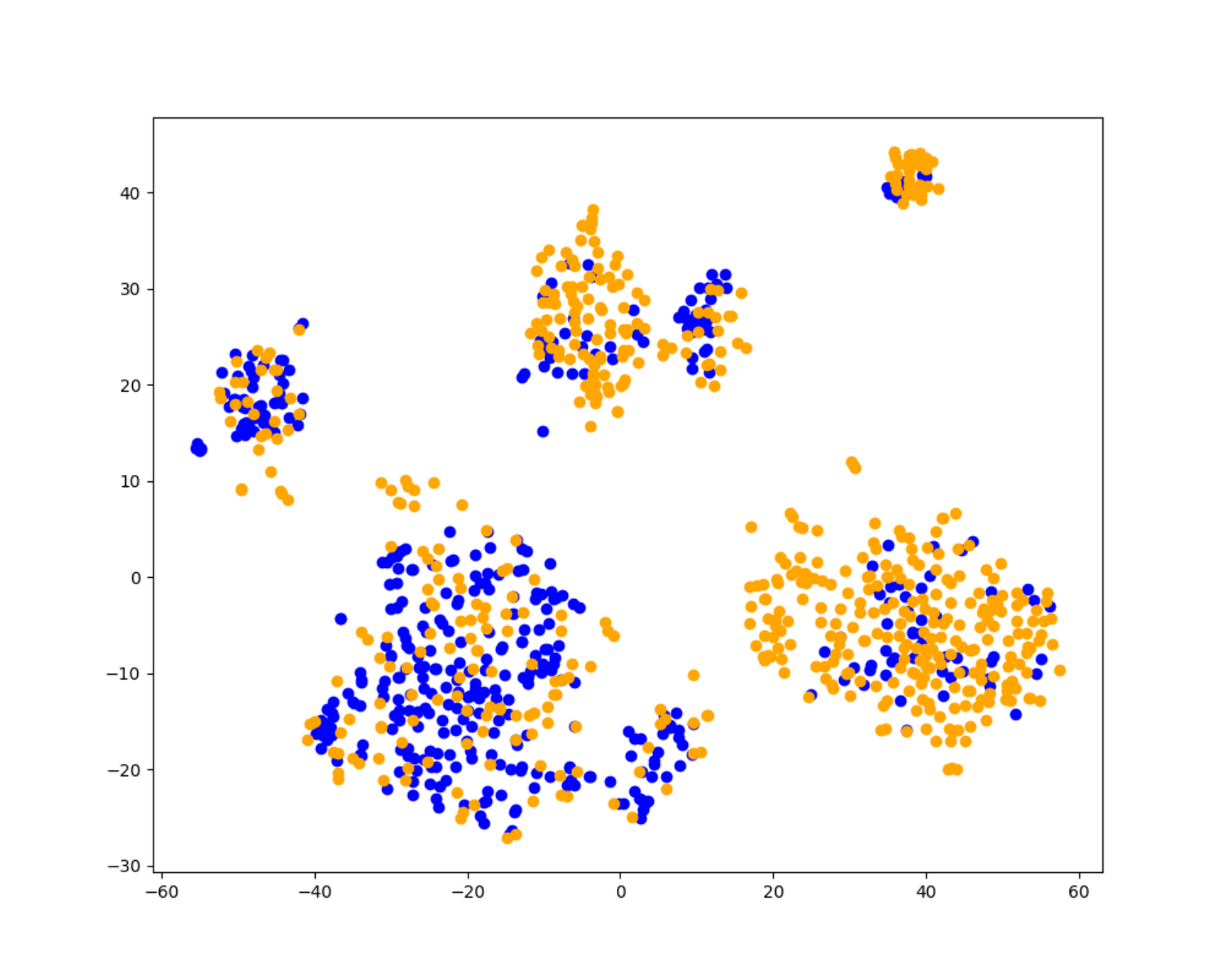}
        \caption{t-SNE on CoQA.}
        \label{fig:prob1_6_2}
    \end{minipage}%
    \begin{minipage}{0.5\textwidth}
        \centering
        \includegraphics[width=0.9\linewidth, height=0.3\textheight]{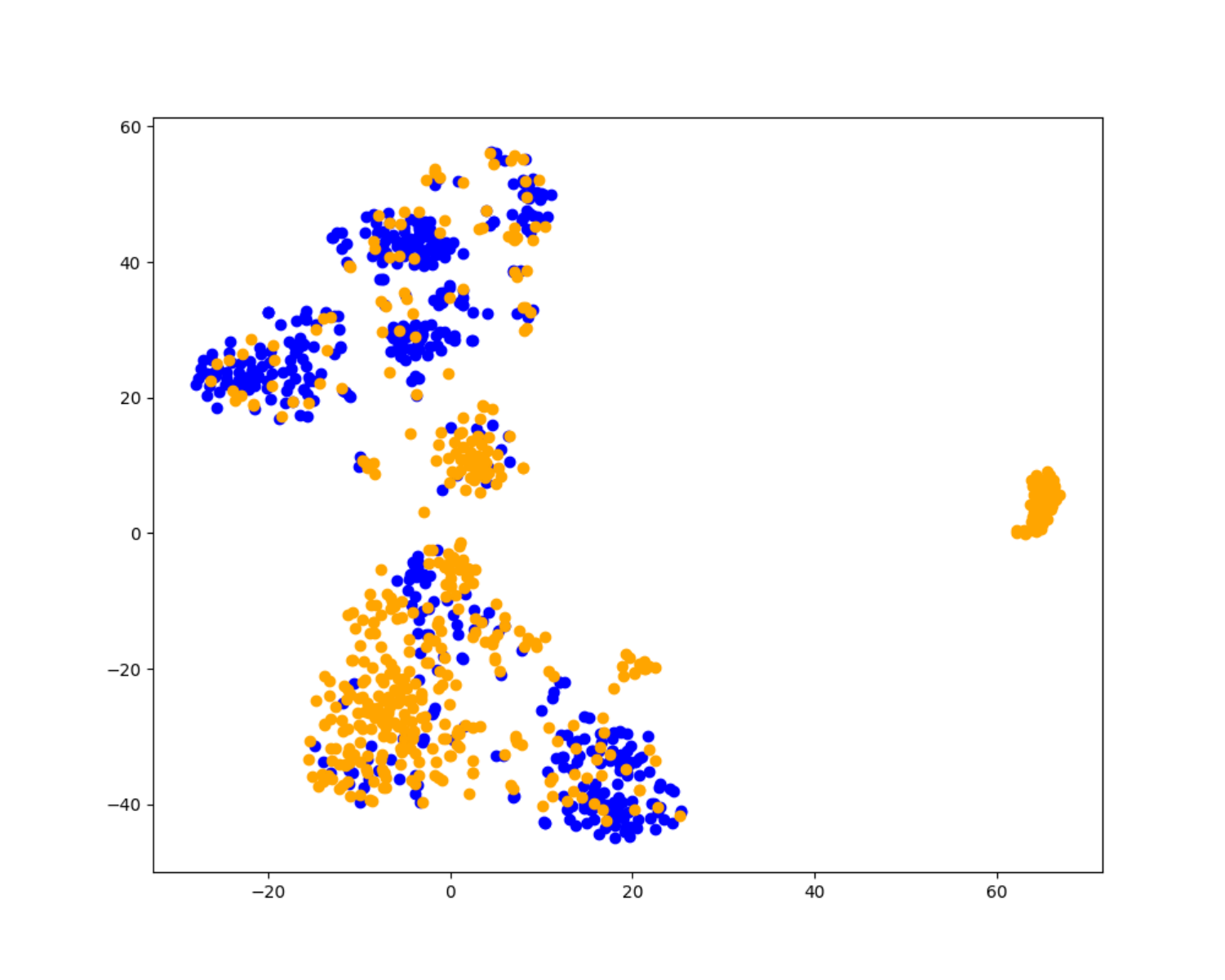}
        \caption{t-SNE on TydiQA.}
        \label{fig:prob1_6_1}
    \end{minipage}
\end{figure}

\begin{figure}[!htb]
    \centering
    \begin{minipage}{.5\textwidth}
        \centering
        \includegraphics[width=0.9\linewidth, height=0.3\textheight]{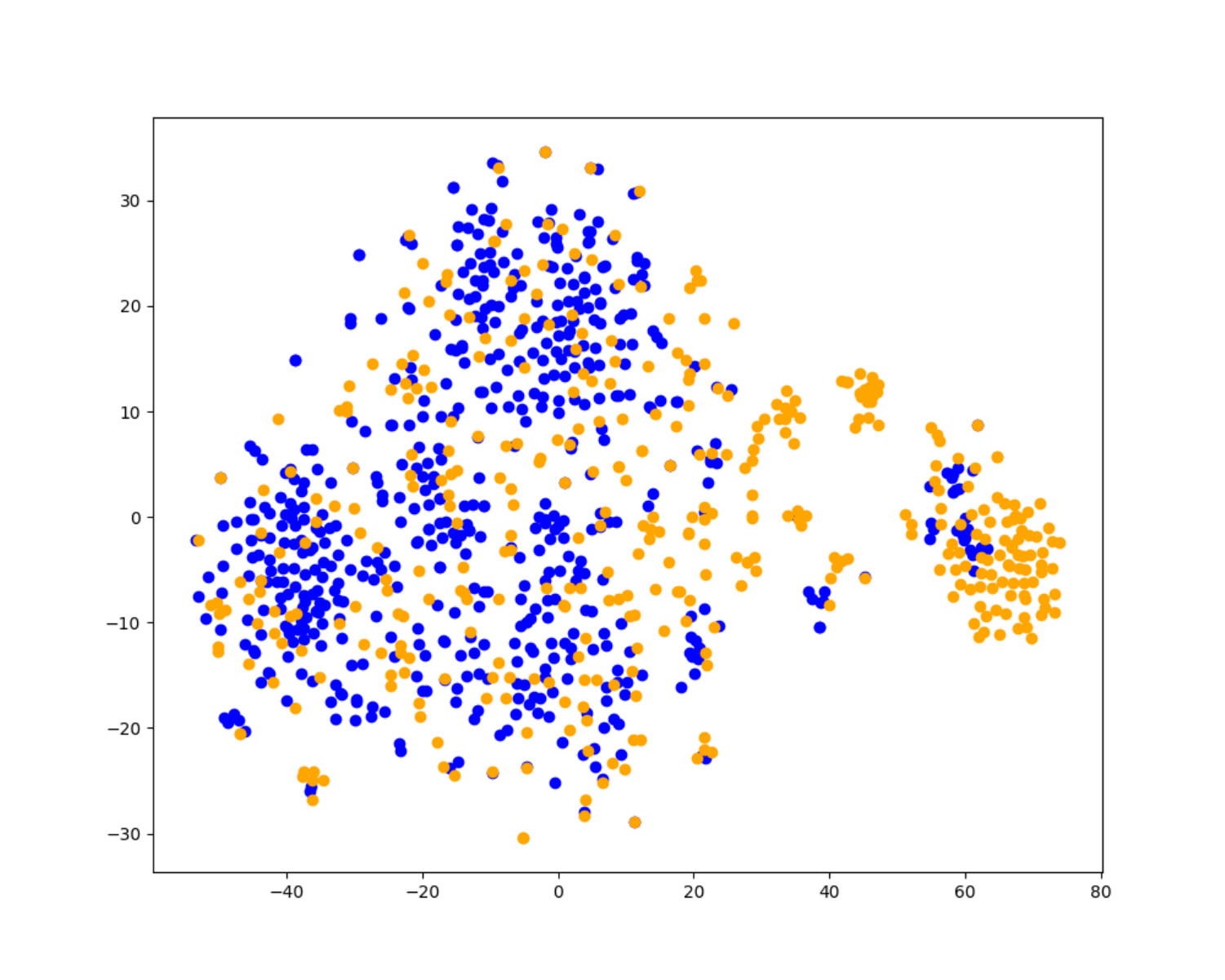}
        \caption{t-SNE on TriviaQA.}
        \label{fig:prob1_6_2}
    \end{minipage}%
    \begin{minipage}{0.5\textwidth}
        \centering
        \includegraphics[width=0.9\linewidth, height=0.3\textheight]{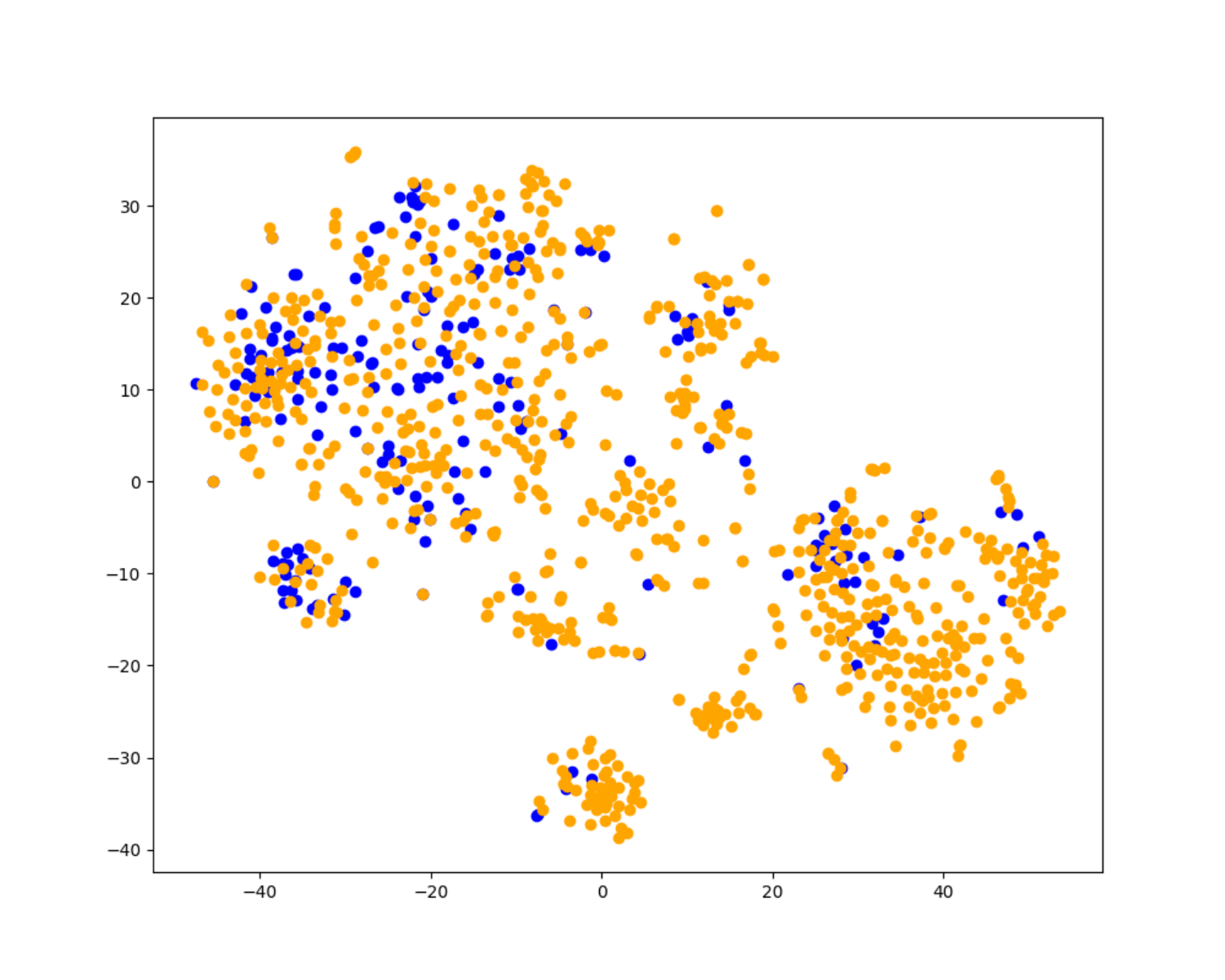}
        \caption{t-SNE on HotpotQA.}
        \label{fig:prob1_6_1}
    \end{minipage}
\end{figure}

\clearpage
\section{Examples for the controlled mixing data}
\label{app:mix}
We show more examples or our synthetic experiments where we ask models to continue a ground-truth answer,
\begin{table}[h]
\renewcommand{\arraystretch}{1.1}
\begin{center}
\small
\begin{tabular}{p{15.55cm}}
\toprule
\texttt{Answer these questions: Q: Who played the title roll in the Flint films? A: }\cr
\midrule
- \textbf{Model output: } \texttt{John Cusack} $\left[\text{7.54, 8.61, 13.34, 5.05}\right]$\cr
- \textbf{Ground-truth: } \texttt{James Coburn} $\left[\text{7.88, 9.96, 4.79}\right]$ \cr
- \textbf{Mixed output: } \texttt{James {\color{blue}Garner}}  $\left[\text{7.87,  8.72,  4.58 , 6.98}\right]$ \cr

\midrule
\texttt{Answer these questions: Q: Which man other than Billy McNeill and Martin O'Neil has managed both Aston Villa and Celtic? A: }\cr
\midrule
- \textbf{Model output: } \texttt{1970s Celtic manager Jock Stein.} \cr
$\left[\text{7.98, 8.05, 8.98, 7.08, 8.76, 10.15, 8.22, 11.30, 9.22, 6.78, 6.76, 8.57, 7.33, 6.87}\right]$\cr
- \textbf{Ground-truth: } \texttt{DR JOSEF VENGLOS} \cr
$\left[\text{8.32, 7.66, 8.11, 8.87, 8.80, 7.74, 8.38, 10.76, 5.74}\right]$ \cr
- \textbf{Mixed output: } \texttt{DR JOSEF  {\color{blue}SMEDA}} \cr
$\left[\text{8.32, 7.66, 8.11, 8.87, 8.80, 7.50, 11.17, 7.46}\right]$ \cr

\midrule
\texttt{Answer these questions: Q: Which famous vehicle has used license plates SCV 00919 and SCV 1 (through to SCV 9)? A: }\cr
\midrule
- \textbf{Model output: } \texttt{007's Aston Martin DB5} \cr
$\left[\text{9.50, 10.02, 11.33, 9.06, 8.82, 10.25, 8.07, 9.75, 6.79, 8.46, 6.70}\right]$\cr
- \textbf{Ground-truth: } \texttt{The Popemobile} \cr
$\left[\text{9.00, 9.23, 9.53, 6.72}\right]$ \cr
- \textbf{Mixed output: } \texttt{The  {\color{blue}Black Pearl}} \cr
$\left[\text{9.00, 8.52, 11.39, 7.04, 7.86}\right]$ \cr

\midrule
\texttt{Answer these questions: Q: In the J. M. Barrie play who was the resourceful butler to the Earl of Loam? A: }\cr
\midrule
- \textbf{Model output: } \texttt{Jeeves} \cr
$\left[\text{8.75, 10.51, 6.43}\right]$\cr
- \textbf{Ground-truth: } \texttt{THE ADMIRABLE CRICHTON} \cr
$\left[\text{6.74, 7.82, 9.03, 9.86, 9.91, 10.19, 10.02, 12.58, 5.41}\right]$ \cr
- \textbf{Mixed output: } \texttt{THE ADMIRABLE {\color{blue}BERTIEl}} \cr
$\left[\text{[6.74, 7.82, 9.03, 9.86, 9.91, 7.39, 8.11, 7.29}\right]$ \cr

\midrule
\texttt{Answer these questions: Q: With what invention do you associate the name of Mr. Whitcomb Judson? A: }\cr
\midrule
- \textbf{Model output: } \texttt{The sewing machine} \cr
$\left[\text{9.46, 9.75, 9.40, 5.55, 7.49}\right]$\cr
- \textbf{Ground-truth: } \texttt{ZIP fastener} \cr
$\left[\text{8.69, 10.32, 11.12, 6.47}\right]$ \cr
- \textbf{Mixed output: } \texttt{ZIP  {\color{blue}PERl}} \cr
$\left[\text{[8.69, 10.32, 8.37, 9.05}\right]$ \cr
\bottomrule\hline
\end{tabular}

\vspace{-6pt}
\end{center}
\caption{Examples of mixing distributions increases LIDs.}
\label{table:mix}
\end{table}


\end{document}